# X-Nav: Learning End-to-End Cross-Embodiment Navigation for Mobile Robots

Haitong Wang, *Student Member IEEE*, Aaron Hao Tan, *Student Member, IEEE*, Angus Fung, *Student Member, IEEE*, and Goldie Nejat, *Member, IEEE*

***Abstract*—Existing navigation methods are primarily designed for specific robot embodiments, limiting their generalizability across diverse robot platforms. In this paper, we introduce X-Nav, a novel framework for end-to-end cross-embodiment navigation where a single unified policy can be deployed across various embodiments for both wheeled and quadrupedal robots. X-Nav consists of two learning stages: 1) multiple expert policies are trained using deep reinforcement learning with privileged observations on a wide range of randomly generated robot embodiments; and 2) a single general policy is distilled from the expert policies via navigation action chunking with transformer (Nav-ACT). The general policy directly maps visual and proprioceptive observations to low-level control commands, enabling generalization to novel robot embodiments. Simulated experiments demonstrated that X-Nav achieved zero-shot transfer to both unseen embodiments and photorealistic environments. A scalability study showed that the performance of X-Nav improves when trained with an increasing number of randomly generated embodiments. An ablation study confirmed the design choices of X-Nav. Furthermore, real-world experiments were conducted to validate the generalizability of X-Nav in real-world environments.**

## I. Introduction

Robot navigation in diverse and challenging environments is crucial for mobile robots to perform tasks such as person search and detection [1]-[3], exploration in unknown environments [4], [5], and robot-guided navigation [6], [7]. However, existing robot navigation methods are dependent on embodiment-specific kinematics, dynamics, and sensory configurations. Embodiment-specific design limits generalization across embodiments, as policies trained for one robot embodiment often cannot be transferred to other robots with different morphological properties [8].

To address this limitation, cross-embodiment navigation methods train a single generalized policy to be deployed on a wide range of robot embodiments of the same (e.g., wheeled robots [9]) or different robot types (e.g., wheeled and legged robots [10]). Namely, existing methods have mainly used either imitation learning (IL) [9]-[16], or deep reinforcement learning

Manuscript received July 10, 2025; Revised September 17, 2025; Accepted October 17, 2025. This paper was recommended for publication by Editor Abhinav Valada upon evaluation of the Associate Editor and Reviewers' comments. This work was supported by the Natural Sciences and Engineering Research Council of Canada (NSERC). *Corresponding author: Haitong Wang.*

The authors are with the Autonomous Systems and Biomechatronics Laboratory (ASBLab), Department of Mechanical and Engineering, University of Toronto, Toronto, ON M5S 3G8, Canada (e-mail: haitong.wang@mail.utoronto.ca; aaronhao.tan@utoronto.ca; angus.fung@mail.utoronto.ca; nejat@mie.utoronto.ca.

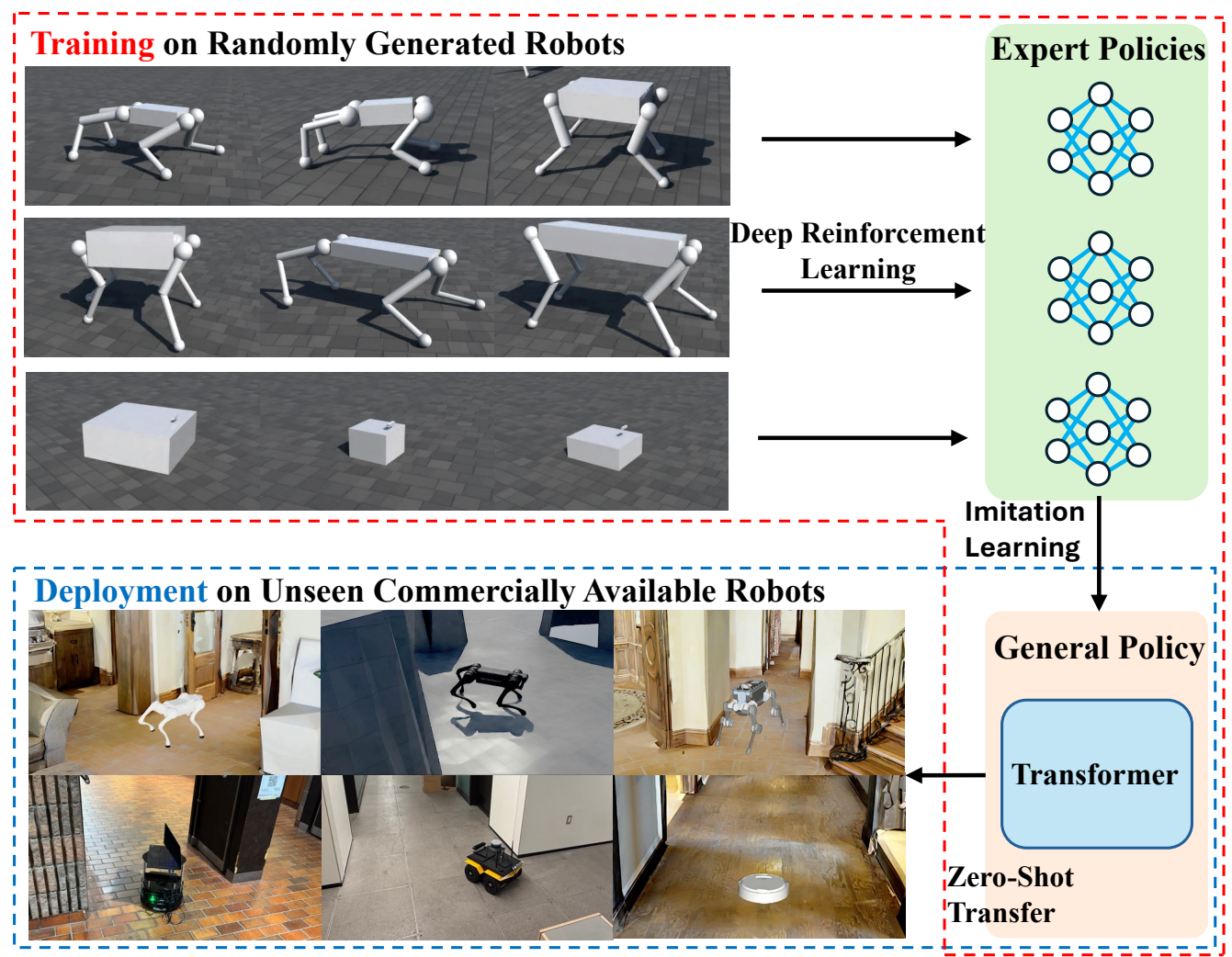

**Fig. 1.** An overview of X-Nav. X-Nav trains a single end-to-end general navigation policy using randomly generated robot embodiments, which achieves zero-shot transfer to a variety of unseen wheeled and quadrupedal robots in simulated and real-world environments.

(DRL) [8], [17]-[22]. In particular, IL methods learn a navigation policy from datasets collected on heterogeneous robot platforms (e.g., wheeled and quadrupedal robots). On-the-other-hand, DRL methods usually learn to navigate by training on diverse existing or randomly generated wheeled or quadrupedal robots. However, these methods often rely on embodiment-specific modules such as waypoint tracking controller [14], or dynamics models to predict future robot poses [17]. Therefore, the parameters of these embodiment-specific modules need to be tuned for each robot embodiment. Moreover, for DRL methods that focus on locomotion, their policies *only learn to track* given velocity commands and they are unable to *plan velocities*. As a result, these methods would require human teleoperation [20] or a separate navigation planner to plan robot velocities [18] to perform navigation.

In this paper, we propose X-Nav, a novel two-stage learning framework for end-to-end cross-embodiment navigation, where a *single* generalized navigation policy is developed that can be deployed on a wide range of mobile robots, Fig. 1. In particular, we focus on wheeled and 12-degree-of-freedom (DOF) quadrupedal robots as they are the most widely used robot mobility types for navigation. In particular, wheeled robots have energy-efficient motion on flat terrains, whereas quadrupeds can traverse across diverse and challenging terrains [23]. Our navigation policy is end-to-end in that it directly maps visual and proprioceptive observations to executable control commands without relying on embodiment-specific models. In this paper, we define executable action commands as the low-level action representations used in navigation and locomotion [15], [24]. For quadrupeds, the actions are desired joint positions [24]; for wheeled robots, the actions are base linear and angular velocities [15].

The key contributions of this work are: 1) the development of a novel end-to-end cross-embodiment navigation approach, which can be deployed on different types of mobile robots, including wheeled and quadrupeds; and 2) the introduction of a two-stage learning framework that integrates both expert policy learning using DRL with privileged observations, and general policy distillation using IL with a transformer model.

## II. Related Works

We discuss the existing literature on cross-embodiment navigation that have used: 1) IL [9]-[16] or 2) DRL [8], [17]-[22].

### A. Imitation Learning-based Methods

IL-based methods learn a cross-embodiment navigation policy by training on datasets collected from heterogeneous robot platforms in real-world [9]-[15] or simulated environments [16]. These methods use robot visual observations (i.e., RGB images) and goal location images as inputs. They then extract spatial features from these input images using visual encoders such as EfficientNet [25] and ResNet [26]. The extracted image features are used to generate relative waypoints [10]-[13] or velocities [9], [15]. This is achieved through the use of fully connected layers (FCLs) [9], [11], [15], transformer blocks, [10], [12]-[14] or diffusion action heads [10], [13], [16]. FCLs directly map the image features to robot actions. Transformer blocks use self-attention layers to account for the spatial-temporal features of robot observations [27]. Diffusion head generates robot actions by progressively denoising action vectors with image features as conditioning [28]. Then, embodiment-specific controllers are used to track the generated waypoints or velocities.

IL methods have been trained on datasets containing robot navigation trajectories collected on various robot platforms. These datasets typically include RGB image observations and robot poses collected in indoor environments (e.g., SACSoN [29]), or outdoor off-road and sidewalk environments (e.g., GNM [11]). Evaluation of the IL methods were conducted in both indoor [9]-[15], [16] and outdoor [9]-[13], [16] real-world environments using different wheeled robots [9]-[14], [16], and/or quadruped robots [10], [15], [16].

### B. Deep Reinforcement Learning-based Methods

DRL-based methods consist of: 1) hierarchical [17], [18], and 2) monolithic methods [8], [19]-[22].

Hierarchical methods integrate high-level planning and low-level control modules. For example, in [17], a model predictive control (MPC) framework consisting of a dynamics module and a perception module was used for wheeled robot navigation in outdoor environments. In [18], a quadrupedal robot navigation system was developed using the SAC+AE [30] DRL method. It consisted of a general high-level and a low-level policy. The high-level policy used a robot embedding network to generate robot-specific embeddings and a multi-layer perceptron (MLP) to generate robot base velocities. The low-level policy used an MPC policy to track robot velocities. It was evaluated in indoor house environments using different quadrupeds.

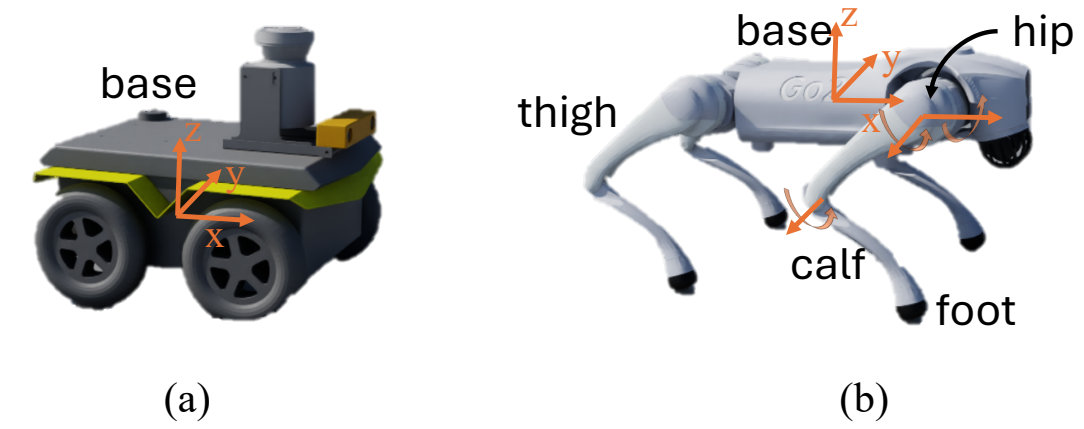

(a) (b)

**Fig. 2.** The base frame of a: (a) wheeled robot, and (b) quadrupedal robot.

Monolithic methods have been used to map observations such as proprioception (e.g., joint positions, joint velocities) [8], and joint descriptions (e.g., torque limits, velocity limits) [19] directly to robot actions. The generated actions are either high-level discrete actions (move forward, turn left) for wheeled robots [22] or low-level continuous actions (joint positions) for quadrupeds [8], [19]-[21]. To learn policies that generalize across different embodiments, these methods were often trained using randomly generated [8], [20]-[22] or existing robot embodiments [19]. Monolithic methods have mainly been trained using Proximal Policy Optimization (PPO) [31]. They have been evaluated in both indoor [8], [21], [22] and outdoor [19], [20] environments using various wheeled robots [22] or quadrupedal robots [8], [19]-[21].

### C. Summary of Limitations

The aforementioned IL-based methods depend on embodiment-specific controllers to track the predicted waypoints [10], [11] or velocities [9], [15]. Each robot embodiment requires individualized tuning of these controllers to ensure accurate execution of navigation commands. DRL-based methods also require embodiment-specific models to either generate low-level control commands [17], predict future robot poses [18], or execute high-level discrete actions [22], which requires embodiment-specific tuning when deployed on new embodiments. Furthermore, DRL methods that have focused only on quadruped locomotion primarily learn to track velocity commands but lack the ability to plan robot velocities [8], [19]-[21]. Therefore, they require manual teleoperation [20] or external velocity planner to perform navigation [18].

To address the above limitations of relying on embodiment-specific modules and the lack of velocity planning, X-Nav provides an end-to-end navigation approach that directly maps robot observations to executable low-level commands. This is achieved through a novel two-stage training framework, where multiple expert policies are trained on randomly generated robots, and distilled into a general navigation policy.

## III. Problem Formulation

The cross-embodiment navigation problem requires learning a single policy $\pi_\theta$ that enables an unseen mobile robot $r \in \mathcal{R}_{wheel} \cup \mathcal{R}_{quad}$ to navigate from a starting position $l_s \in \mathbb{R}^2$ to a given goal position $l_g \in \mathbb{R}^2$ at a reference speed $v_{\text{ref}} \in \mathbb{R}$ in an unknown cluttered environment. The robot navigation is guided by: 1) the visual observations which are represented by depth images $o_{depth} \in \mathbb{R}^{H \times W}$ from an onboard depth camera,

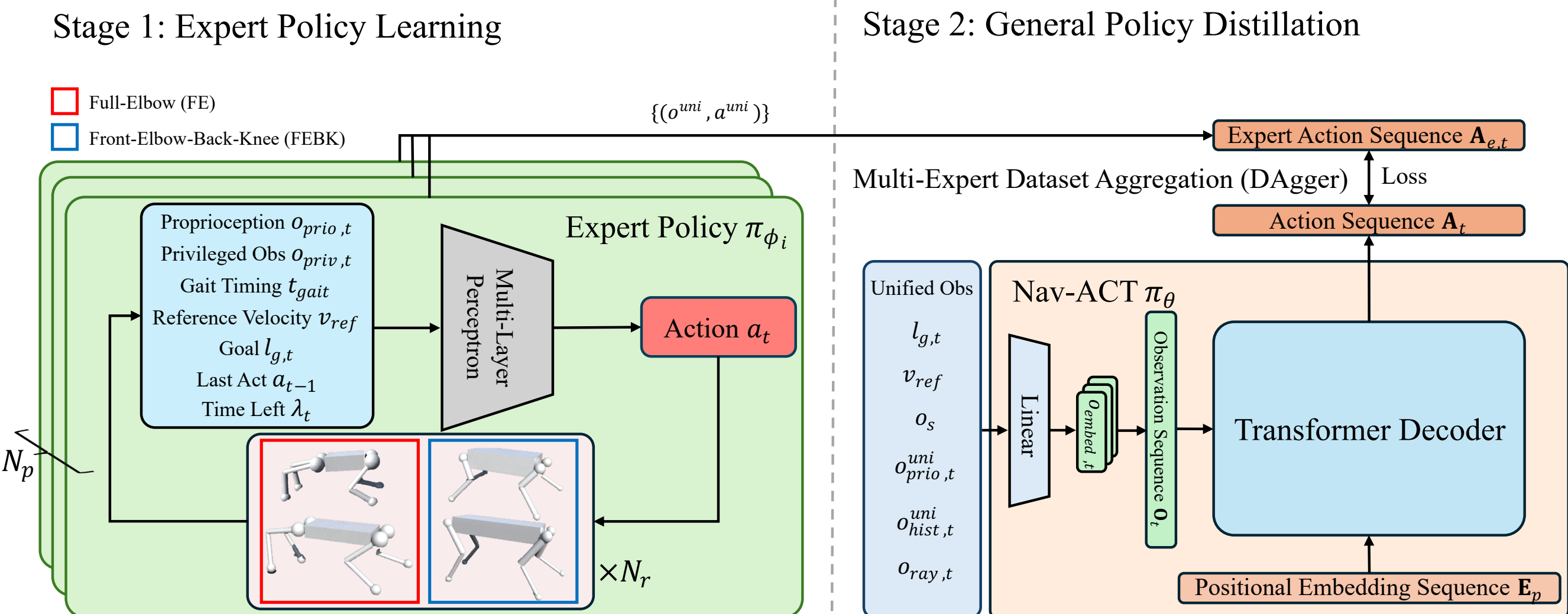

**Fig. 3.** The proposed X-Nav framework consists of two stages: 1) Expert Policy Learning, and 2) General Policy Distillation. In Stage 1, multiple expert policies are trained on randomly generated robot embodiments using DRL with privileged observations. In Stage 2, the knowledge of expert policies is distilled into a single general policy using IL.

2) the 2D goal position $l_g$, 3) the reference speed $v_{ref}$, and 4) the proprioceptive observations $o_{prio}$ obtained from an inertial measurement unit (IMU) and motor encoders. Based on these observations, the robot executes an action $a$ at each timestep $t$. The objective is to minimize its travel distance:

$$\min \mathbb{E}\left[d_{nav}(l_s, l_g)\right], \tag{1}$$

where $d_{nav}(\cdot,\cdot)$ is a function representing travel distance.

We consider both wheeled and quadruped robots. For a wheeled robot $r \in \mathcal{R}_{wheel}$, the proprioception $o_{prio}^{wheel} \in \mathbb{R}^2$ is a 2D vector representing robot linear and angular velocity in the robot base frame (Fig. 2 (a)), and action $a^{wheel} \in \mathbb{R}^2$ is a 2D vector representing the desired linear and angular velocity. For a quadruped $r \in \mathcal{R}_{quad}$, the proprioception $o_{prio}^{quad} \in \mathbb{R}^{30}$ is a 30D vector representing the concatenation of the robot base velocity $o_{prio,vel}^{quad} \in \mathbb{R}^3$, the gravity projected in the robot base frame $o_{prio,g}^{quad} \in \mathbb{R}^3$, the positions of the 12 joints (i.e., hip, thigh and knee joints) $o_{prio,jp}^{quad} \in \mathbb{R}^{12}$ and the corresponding joint velocities $o_{prio,jv}^{quad} \in \mathbb{R}^{12}$. The robot base frame is defined with its origin at the center of the robot trunk, Fig. 2 (b). The action $a^{quad} \in \mathbb{R}^{12}$ represent the 12 desired joint positions. A Proportional-Derivative (PD) controller is used to track the desired position for each joint of a quadruped robot [24]:

$$\tau = K_p(a^{quad} - q^{quad}) - K_d \dot{q}^{quad}, \tag{2}$$

where $\tau$ is the motor torque, $K_p$, $K_d$ are the PD gains, and $q^{quad}$ represents the actual joint position.

## IV. X-Nav Architecture

The proposed X-Nav architecture, shown in Fig. 3, consists of two stages: 1) Expert Policy Learning, and 2) General Policy Distillation. In Stage 1, randomized robot embodiments are used to train expert policies using DRL with privileged observations. In Stage 2, a general policy was distilled from the multiple expert policies using a transformer model and IL.

### *A. Stage 1: Expert Policy Learning*

In Stage 1, we train $N_p$ expert policies $\pi_{\phi_i}$ $(i = 1, \dots, N_p)$ in simulation using DRL. Each expert policy is trained on $N_r$ different robot embodiments of a same type.

#### *1) Observation and Action Space*

At each timestep $t$ $(0 \leq t < T)$, the policy observation is $o_t = \left[a_{t-1}, o_{\text{prio},t}, v_{ref}, l_{g,t}, \lambda_t, o_{\text{priv},t}\right]$, where $a_{t-1}$ denotes the last robot action at timestep $t-1$, $\lambda_t = 1 - \frac{t}{T}$ denotes the normalized time remaining in the episode, $T$ is the maximum timesteps of the episode. $o_{priv,t} = \left[o_{embod}, o_{scan,t}\right]$ denotes the privileged observations which include embodiment parameters $o_{embod}$ and terrain height scans $o_{scan,t}$. For wheeled robots, embodiment parameters include the robot mass and its 3D size. For quadrupeds, embodiment parameters include base size, base mass, thigh size, thigh mass, calf size, calf mass, and motor PD gains. $o_{scan}$ represents the terrain height of a local region surrounding the robot. The actions of the expert policies (i.e., $a^{wheel}$ or $a^{quad}$) are defined above in Section III. Quadruped policies have an additional observation of gait timing $t_{gait}$ [32] that provides timing reference for learning the trotting gait.

#### *2) Embodiment Randomization*

Embodiment randomization is implemented to ensure that real-world robot embodiments are within the distribution of the randomly generated robots used during training [8]. We generate random wheeled and quadrupedal robots by sampling $o_{embod}$. To ensure that the randomly generated quadrupeds can generate sufficient motor torques, we design a robot template and utilize it to compute motor PD gains [8]:

$$K_{p/d} = v_{temp} \times \frac{m_{gen}}{m_{temp}} \times v, \tag{3}$$

where $v$ denotes the randomly sampled value, $v_{temp}$ denotes the PD gains of the robot template, $m_{gen}$ denotes the total mass of the generated robot, and $m_{temp}$ denotes the total mass of the robot template. Furthermore, we consider two leg configurations for the generated quadruped embodiments, Fig. 3: Full-Elbow (FE), and Front-Elbow Back-Knee (FEBK) as they are the most commonly used configurations [8].

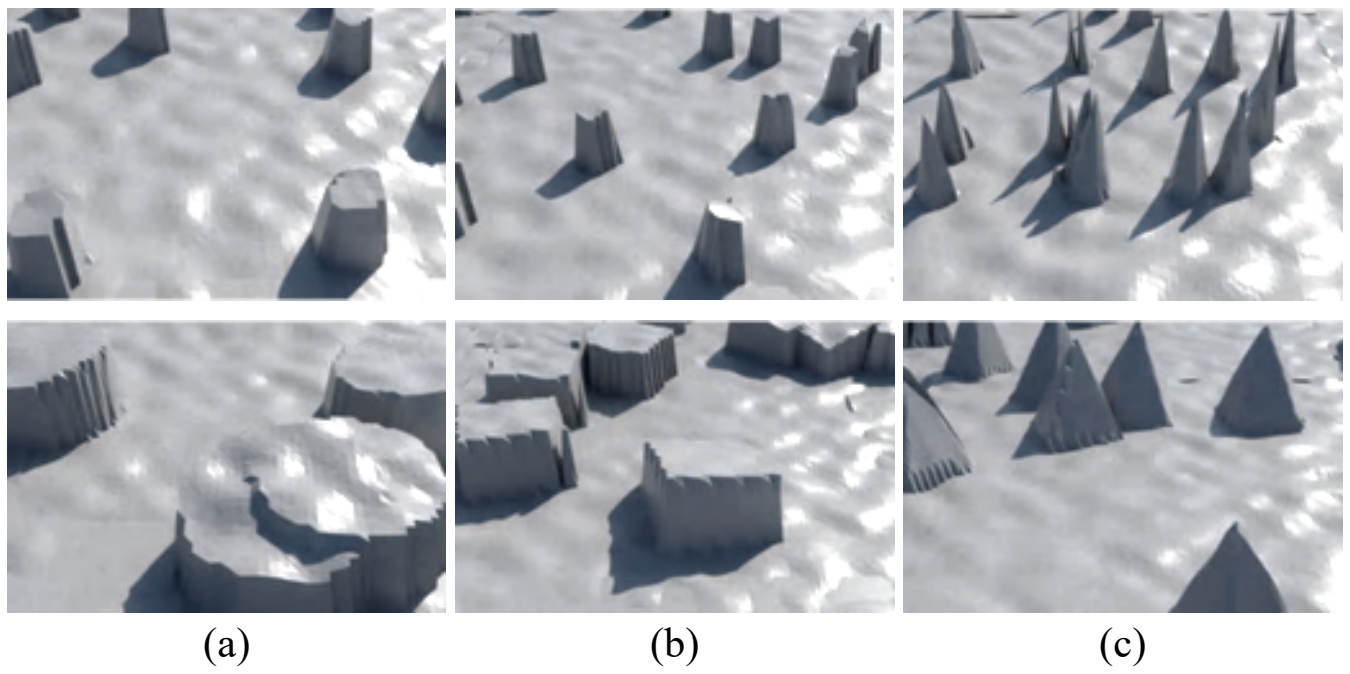

(a) (b) (c)

**Fig. 4.** The obstacles used for expert policy learning. (a) cylinders, (b) boxes, (c) pyramids.

*3) Deep Reinforcement Learning*

We utilized the PPO DRL algorithm [31] to train the expert policies due to its stability and efficiency in learning robot locomotion and navigation [24], [33]. Each policy network $\pi_{\phi_i}$ was implemented as an MLP. We perform massively parallel training [24] by simultaneously training on $N_r$ robots.

*4) Rewards*

The reward function, $r$, has been designed to encourage mobile robots to perform obstacle-free navigation and promote smooth and efficient movements. $r$ includes positive task rewards $r_{task}$, and negative regularization rewards $r_{reg}$. The final reward is computed as $r = r_{task} * exp(c_{reg} * r_{reg})$ [32]. Wheeled and quadrupedal robots share the same task rewards but different regularization rewards $r_{reg}^{wheel}$, and $r_{reg}^{quad}$. The detailed explanation of the reward function is in Table A.I in the supplementary material: https://cross-embodiment-nav.github.io.

*5) Curriculum Learning*

We use a game-inspired curriculum [24] to progressively train the navigation policy from simple to complex environments. The overall environment consists of $n_x \times n_y$ subfields. Each subfield contains randomly generated obstacles or terrains. The obstacles consist of the three types: box, pyramid, and cylinder, Fig. 4. These fundamental geometric primitives are used as approximations of common real-world structures such as walls, beds, and desks. Each obstacle type has difficulty levels represented by its density and the obstacle size. For quadruped robots, we additionally include challenging terrains that are not traversable by wheeled robots, which include stairs, randomized grid terrains, and rough terrains with large elevation variations, Fig. 5. Training starts at the lowest difficulty level. The progression of robots through the difficulty levels is governed by two distance thresholds $\sigma_{close}$ and $\sigma_{far}$ ($\sigma_{close} < \sigma_{far}$). When a robot reaches its goal position at the end of an episode ($d_{goal} < \sigma_{close}$), the robot is promoted to a higher difficulty level. Conversely, if the goal distance exceeds $\sigma_{far}$ ($d_{goal} > \sigma_{far}$), the robot is demoted to a lower difficulty level.

*6) Domain Randomization*

We use domain randomization [34] to improve robustness of the expert policies for sim-to-real transfer. First, friction coefficients and robot base mass are sampled using a uniform distribution from predefined ranges at the beginning of each episode. Second, push disturbances are simulated for quadruped training by adding a randomly sampled velocity to the robot base every $T_{push}$ seconds. Third, random noise is added to the proprioceptive observations $o_{prio}$ and the actions $a$. Fourth, random motor delays are added to the action execution.

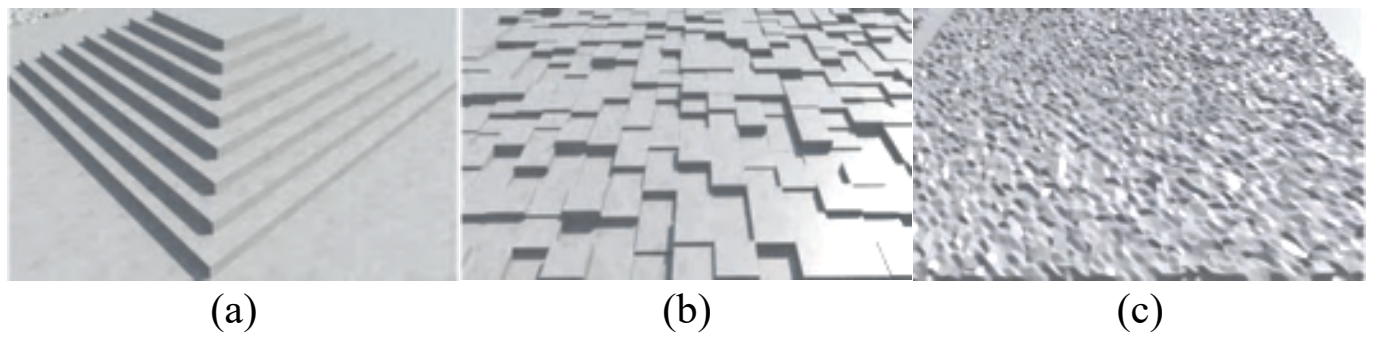

(a) (b) (c)

**Fig. 5.** The challenging terrains for quadrupeds: (a) stairs, (b) random grid, and (c) random rough terrains with large variations in elevation.

*B. Stage 2: General Policy Distillation*

We distill the trained expert policies into a single general policy $\pi_\theta$ using imitation learning.

*1) Unified Observation and Action Space*

The unified observation $o_t^{uni}$ at timestep $t$ is:

$$o_t^{uni} = \left[l_{g,t}, v_{ref}, o_s, o_{prio,t}^{uni}, o_{hist,t}^{uni}, o_{ray,t}\right], \quad (4)$$

where $o_s \in \mathbb{R}^2$ is the robot base length and width, and $o_{prio,t}^{uni}$ is the unified proprioception. For wheeled robots, where joint position and velocity data are absent from proprioception $o_{prio}^{wheel}$, we pad the dimensions with zeros to match $o_{prio}^{quad}$. $o_{hist,t}^{uni}$ is the concatenation of the last $N_{hist}$ frames of $a^{uni}$ and $o_{prio}^{uni}$. $o_{ray,t}$ is the unified laser rays for distance measurement derived from the raw depth image $o_{depth}$. We project depth pixels of $o_{depth}$ to a horizontal plane and extract distances at evenly spaced angles. The resulting laser scan is then interpolated into a unified format with a field of view (FOV) of $\theta_{fov}$. This ensures consistency across different cameras.

The unified action, $a^{uni} \in \mathbb{R}^{14}$, is a 14D vector with the first two dimensions representing the linear and angular velocity of wheeled robots, and the last 12 dimensions representing the target joint positions for quadrupeds.

*2) Navigation Action Chunking with Transformer*

We introduce Navigation Action Chunking with Transformer (Nav-ACT), a transformer model to generate $a^{uni}$ from $o_t^{uni}$ for cross-embodiment navigation, Fig. 3. Nav-ACT consists of: 1) an encoder to convert the unified observations $o^{uni}$ into embeddings, and 2) a transformer decoder to generate an action chunk conditioned on the observation sequence.

At each timestep $t$, the unified observation $o_t^{uni}$ is converted into an embedding $o_{embed,t}$ using a linear encoder. $S_o$ consecutive steps of observation embeddings $o_{embed}$ are stacked to construct an observation sequence $\mathbf{O}_t = \left[o_{embed,t-S_o+1}, \dots, o_{embed,t}\right]$. The transformer decoder takes as input a sequence of $S_a$ learnable positional embeddings $E_p$. It then uses the observation embedding sequence $\mathbf{O}_t$ for cross-attention computation to generate an action sequence $\mathbf{A}_t$, $\mathbf{A}_t = \left[a_{t-S_a+1}, \dots, a_t\right]$, where $S_a$ denotes the sequence length. $\mathbf{A}_t$ is generated in a single forward pass for computation efficiency.

*3) Multi-Expert Dataset Aggregation with Action Chunking*

We use the Dataset Aggregation (DAgger) [35] IL method to distill the multiple expert policies into the single Nav-ACT model. DAgger is used as it mitigates compounding errors caused by distributional shift and has been widely applied in robot locomotion and navigation [36]. All the randomly generated embodiments interact with the environment in simulation using Nav-ACT. Namely, at each timestep $t$, Nav-ACT generates an action sequence $\mathbf{A}_t$ and the first action from the sequence, $a_t = \mathbf{A}_t[0]$, is executed. At the same time, for each robot embodiment, the corresponding expert policy $\pi_\phi$ is queried to generate an expert action $a_{e,t}$. To enable action chunking, the expert actions are stacked into action chunks $\mathbf{A}_{e,t} = [a_{e,t-S_a+1}, \ldots, a_{e,t}]$. Nav-ACT is trained using the mean squared error (MSE) loss:

$$\mathcal{L}_{Nav-ACT} = \frac{1}{S_a}\sum_{i=1}^{S_a} \|\mathbf{A}[i] - \mathbf{A}_e[i]\|^2. \quad (5)$$

*4) Inference*

At inference time, for: 1) wheeled robots, we use temporal ensemble (TE) to improve smoothness and avoid jerky movements, and 2) for quadrupeds, we disable TE and only take the first action from the action sequence $\mathbf{A}_t$ to ensure real-time adaptation to dynamic changes during navigation. Namely, for wheeled robots, at each timestep $t$, Nav-ACT generates action predictions $\mathbf{A}_t$, then the last $S_a$ predictions $\mathbf{A}_i$, $i = (t - S_a + 1, \ldots, t)$ are taken to compute the action prediction for the current timestep $t$ from each $\mathbf{A}_i$. Weighted average is applied to generate the action:

$$a_t = \frac{\sum_{i=t-S_a+1}^{t} w_i \mathbf{A}_i[t-i]}{\sum_{i=t-S_a+1}^{t} w_i}, \quad (6)$$

where $w_i = exp(-k * (i - t + S_a - 1))$ and $k$ is a positive constant. For quadrupeds, at each timestep, we take the first action from the latest action sequence, $a_t = \mathbf{A}_t[0]$.

## V. Policy Training

### A. Expert Policy Learning

*1) Setup:* We trained $N_p = 5$ expert policies using DRL. Three of them were trained in the environments with randomly generated obstacles: one for small-sized quadrupeds (i.e., mass < 30kg), one for large-sized quadrupeds (i.e., mass > 30kg), and one for wheeled robots. Another two expert policies were trained in the environments with challenging terrains: one for small-sized quadrupeds and one for large-sized quadrupeds. Quadrupeds were split into small and large categories due to the significant differences in their dynamics (e.g., inertia, torque) to reduce variability and promote stable training. For expert policy, $N_r = 4096$ robot embodiments were generated by sampling parameters from Table B.I in the supplementary material, and domain randomization followed the ranges defined in Table C.I in the supplementary material. For curriculum learning, the overall training environment consisted of 384 subfields arranged in 6 rows and 64 columns, with each subfield measuring $10 \times 10$ m. The other hyperparameters are defined in Table D.I in the supplementary material. The values of the hyperparameters were empirically determined during training. Each policy network has (1024, 512, 256) units. The Isaac Sim simulator and Isaac Lab framework [37] were used.

*2) Training*: All training was done on a workstation with an NVIDIA RTX 4090 GPU, an Intel Core i9-13900KF CPU and 32GB RAM. Each expert policy was trained with a batch size of 24576 for 4000 epochs. The Adam optimizer [38] was used with a learning rate of 0.001.

### B. General Policy Distillation

*1) Setup:* For observation, we used $N_{hist}$ of 5, $\theta_{fov}$ of 90°, the minimum and maximum ray distance of 0.2 m and 8 m, and number of laser rays of 128. We set $S_a$ as 6, and $S_o$ as 4. Nav-ACT has 2 transformer layers and 4 heads with an embedding size of 256.

*2) Training:* Nav-ACT was trained with a batch size of 28672, learning rate of 0.0001 for 5000 iterations. The Adam optimizer [38] was used with weight decay of 0.001.

## VI. Simulated Experiments

We evaluated the performance of X-Nav by conducting: 1) a comparison study with state-of-the-art (SOTA) methods to assess the generalizability of X-Nav on unseen robot embodiments, 2) a scalability study to explore X-Nav's performance when trained with increasing numbers of random robot embodiments, and 3) an ablation study to investigate the design choices of X-Nav.

### A. Comparison Study with SOTA

The objective of this comparison study is to evaluate whether X-Nav can generalize to unseen robot embodiments and to compare its performance with existing SOTA methods. We used commercially available robot platforms that were unseen during the training of X-Nav. For wheeled robots, the Clearpath Jackal, Clearpath Dingo, and iRobot Create3 were used. For the quadrupeds, the Unitree A1, Unitree Go2, and ANYmal B were used. Each robot was equipped with a front-facing depth camera with a FOV of 90°. The performance metrics utilized were the: 1) success rate (SR), and 2) the success weighted by the normalized inverse path length (SPL) [39] for measuring the trajectory efficiency. A hundred unseen in-distribution environments were randomly generated using the same obstacles in Section IV.A.5. Quadrupeds were tested on rough terrain with obstacles and the challenging terrains, and wheeled robots were tested on flat terrain with obstacles.

*1) Comparison Methods*: We compared our X-Nav method with the following SOTA methods.

**Visual Navigation Transformer (ViNT) [12]**: The ViNT method uses two visual encoders, and a transformer decoder. It uses visual observations, the goal image and a short history of past observations as input to generate a sequence of waypoints.

**Navigation Diffusion Policy (NavDP) [16]:** The NavDP method uses two visual encoders, a transformer model and a diffusion policy head to generate waypoints from the robot observations of RGB-D images and the goal position. NavDP is selected as it is a SOTA cross-embodiment navigation method that has been applied on various robot embodiments.

TABLE I: COMPARISON STUDY WITH SOTA METHODS

| Method | Go2 | | A1 | | ANYmal B | | Jackal | | Dingo | | Create3 | |
|---|---|---|---|---|---|---|---|---|---|---|---|---|
| | SR | SPL | SR | SPL | SR | SPL | SR | SPL | SR | SPL | SR | SPL |
| ViNT | 45.3 | 0.41 | 58.6 | 0.48 | 35.3 | 0.30 | 65.7 | 0.59 | 56.8 | 0.48 | 79.9 | 0.74 |
| NavDP | 53.1 | 0.48 | 57.8 | 0.52 | 37.2 | 0.29 | 70.2 | 0.64 | 71.0 | 0.63 | 82.4 | 0.73 |
| **X-Nav** | 69.5 | 0.61 | 73.1 | 0.62 | 60.5 | 0.53 | 88.3 | 0.82 | 84.0 | 0.78 | 91.7 | 0.85 |

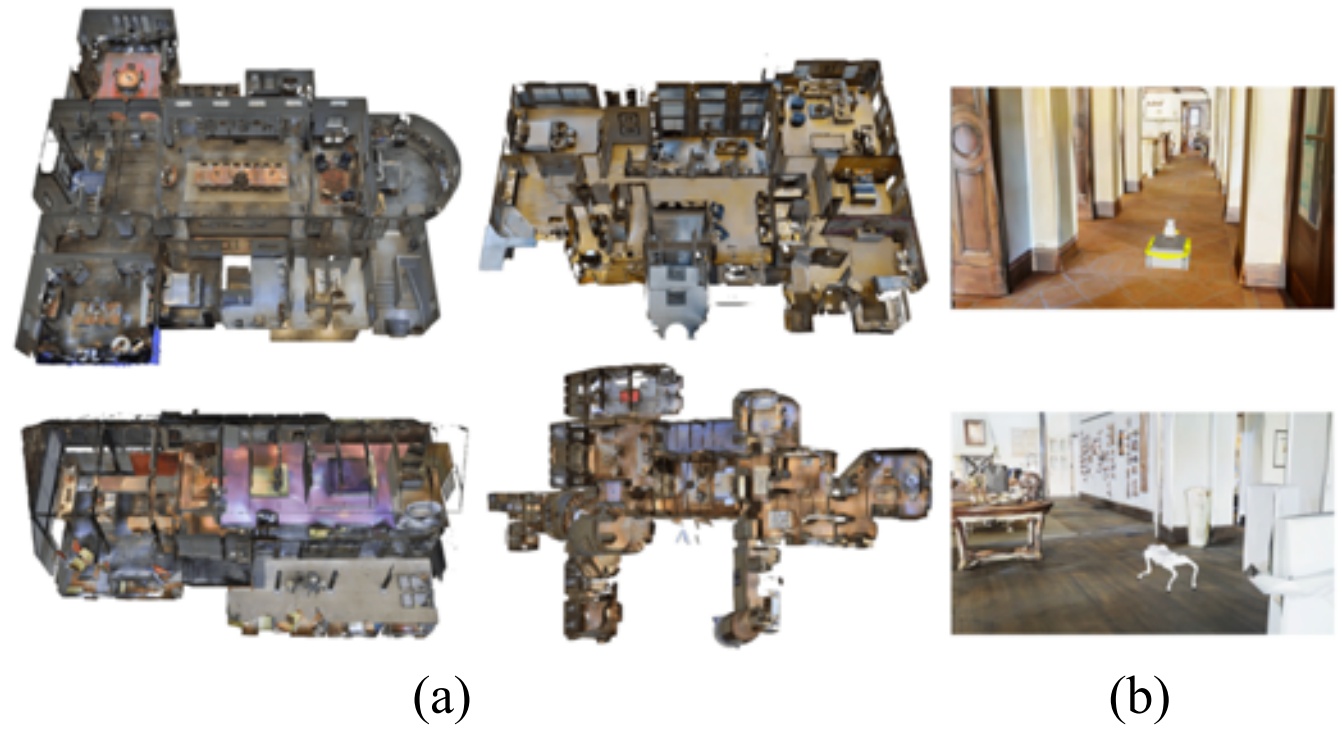

**Fig. 6.** (a) Top view of photorealistic indoor environments from MP3D dataset, (b) Jackal and Go2 deployed inside the environments.

For both ViNT and NavDP, we use a PD controller to convert the generated waypoints into robot linear and angular velocities. These velocities can be directly executed by the wheeled robots. For quadrupeds, we use MorAL [20] to convert the robot velocity commands into target joint positions. We use the open-source model weights for ViNT and NavDP, and train MorAL from scratch using the randomly generated small-sized and large-sized quadrupeds. MorAL is used here as it is a SOTA cross-embodiment locomotion controller for quadrupeds.

*2) Procedure:* We conducted 1000 trials for each method and robot embodiment. At the beginning of each trial, both robot starting and goal positions were randomly placed. The reference speed was randomly sampled. A robot successfully achieved the goal when the distance between its current and goal position was within 0.5m at the end of a trial. Each timestep was 0.02s. A trial terminated when the total timesteps exceeded 1500.

*3) Results:* The SR, SPL of X-Nav and comparison methods are presented in Table I. X-Nav achieved the highest SR and SPL compared to the SOTA methods. ViNT had the lowest performance, as it does not use robot proprioception as input. Thus, it lacks the ability to adapt to robots with different embodiment parameters such as size and mass, which can result in degraded action generation. NavDP also had lower performance than X-Nav. Since it was trained in simulation on data collected from a wheeled robot with a fixed radius, this resulted in overfitting to this particular robot size, limiting its generalizability to adapt to robot embodiments with different body dimensions. In addition, NavDP was trained only on flat terrains and thus, was not able to handle challenging terrains such as stairs. Both NavDP and ViNT rely on intermediate modules, including the PD controller and MorAL, to track generated waypoints. Errors in the predicted waypoints accumulate with the tracking errors of the PD controller and MorAL, leading to navigation failures. In contrast, X-Nav uses an end-to-end policy that directly maps observations to the low-level actions, avoiding error accumulation across modules.

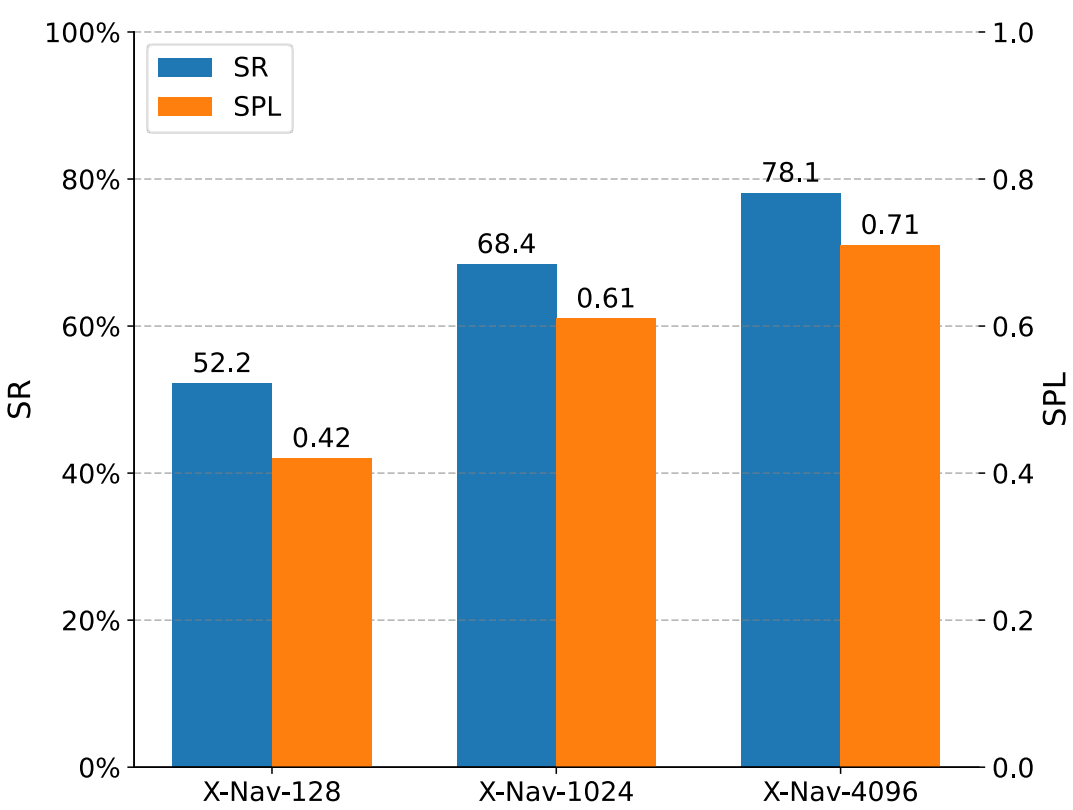

**Fig. 7.** SR and SPL of three X-Nav configurations, using 128, 1024, and 4096 robot embodiments for general policy distillation.

## B. Scalability Study

The objective of this study is to investigate if X-Nav's performance improves with an increasing number of randomized robot embodiments used during training. Three configurations of X-Nav were tested, denoted as X-Nav-128, X-Nav-1024, and X-Nav-4096, where the number indicates the total number of randomly generated robot embodiments used by the general policy distillation stage. For evaluation environments, we used the Matterport3D (MP3D) dataset [40] which contains 3D meshes of real-world indoor home environments. This dataset was selected as it is widely used for robot indoor navigation [41]. In particular, four houses with an average size of $22 \times 25$ m from MP3D were imported into Isaac Sim, Fig. 6. In each house, we randomly generated 20 pairs of start and goal positions and conducted 500 trials for each of the six robot embodiments used in Section VI.A.

*1) Results:* Fig. 7 presents the SR and SPL for all configurations. Overall, X-Nav's SR and SPL increased with the number of random embodiments used for training. This is due to X-Nav being exposed to a broader range of embodiment parameters during training, which allowed it to implicitly infer the embodiment parameters of unseen robots with higher accuracy. As a result, X-Nav was able to generate adaptive navigation actions that accounted for the robot's morphology and dynamics, leading to improved navigation performance. Furthermore, these results demonstrate X-Nav's capability to achieve zero-shot transfer to out-of-distribution photorealistic environments. The average SR and SPL of X-Nav-4096 are comparable to its performance in in-distribution environments from Section VI.A. This is due to the unified ray-based distance representation as input, which allows X-Nav to generalize across different camera configurations. X-Nav was trained with obstacles of varying placement, density, and size to expose the policy to diverse obstacle interactions, enabling generalization to out-of-distribution environments.

## C. Ablation Study

We conducted an ablation study with different variants of X-Nav to evaluate the impact of our design choices on the loss function and inference strategy. These included:

**X-Nav with L1 Loss (w L1):** L1 loss was used for training.

TABLE II: ABLATION STUDY

| Variants | Go2 | | Jackal | |
|---|---|---|---|---|
| | SR | SPL | SR | SPL |
| **X-Nav w L1** | 54.2 | 0.41 | 85.7 | 0.77 |
| **Offline BC** | 65.3 | 0.57 | 86.3 | 0.80 |
| **X-Nav w EC** | 40.5 | 0.37 | 87.2 | 0.79 |
| **X-Nav w/o TE** | 69.5 | 0.61 | 76.3 | 0.68 |
| **X-Nav w TE** | 43.3 | 0.36 | 88.3 | 0.82 |
| **X-Nav** | 69.5 | 0.61 | 88.3 | 0.82 |

**Offline BC**: It was trained on demonstrations collected from the expert policies using the same architecture as Nav-ACT. Each expert policy generates 4096 demonstrations.

**X-Nav with Executing Chunk (w EC):** It executed the entire action chunk $\mathbf{A}_t$ before predicting the next action chunk $\mathbf{A}_{t+s_a}$.

**X-Nav without Temporal Ensemble (w/o TE):** It generated an action sequence $\mathbf{A}_t$ at each timestep and always executed the first action from the current predicted sequence without TE.

**X-Nav with Temporal Ensemble (w TE):** This variant applied TE to both wheeled and quadrupedal robots at inference time.

We conducted 3000 trials for each of the above variants using the Unitree Go2 and Clearpath Jackal platforms. We used the same test environments as in Section VI.A.

*1) Results:* The results are shown in Table II. Overall, X-Nav achieved the highest SR and SPL among all the variants. X-Nav w L1 achieved lower SR and SPL than X-Nav, as the L1 loss penalizes less severely for large action prediction errors compared to the L2 loss, which led to less precise navigation actions. Offline BC also achieved lower performance than X-Nav, as X-Nav using DAgger incorporates online learning with corrective actions. This helped with action generation in scenarios where the robot encounters out-of-distribution states not covered by the offline dataset. X-Nav w EC and X-Nav w TE had the lowest SR and SPL values for quadrupeds. Namely, X-Nav w EC delayed robot adaptation to dynamic changes of robot pose and terrain as it required the execution of an entire action chunk. X-Nav w TE reduced robot responsiveness as it smoothed actions by averaging past predictions, which led to navigation failures. This effect is specific to quadrupeds as their actions are actuator level joint positions that require real-time adaptation to maintain balance. In contrast, wheeled robot actions are linear and angular velocities, where smoothing provides more stable and consistent control that helps reduce abrupt velocity and orientation changes. Accordingly, X-Nav w/o TE achieved the lowest SR and SPL for wheeled robots, highlighting the importance of TE for wheeled robots.

## VII. REAL-WORLD EXPERIMENTS

We conducted real-world experiments in both indoor environments including a hallway, a doorway, a corridor, and outdoor environments including a park with uneven terrain, Fig. 8. To validate X-Nav's effectiveness across different robot embodiments, we used two distinct wheeled robots: the TurtleBot2 and Clearpath Jackal. They differ in physical size, mass, kinematics, and sensor configurations. The TurtleBot2 was equipped with a Kinect sensor, and the Jackal robot was equipped with a ZED 2 stereo camera. Both robots used the Robot Operating System (ROS) Noetic with X-Nav running at 50 Hz. No additional training was done for the real-world deployment. The TurtleBot2 robot navigated a 21 × 8 m indoor environment with a hallway with obstacles such as door mullions and garbage bins, Fig. 8 (a), and a 19 × 5 m indoor environment that included a narrow doorway, Fig. 8 (b). The Jackal robot navigated in a 23 × 7 m indoor narrow corridor with obstacles, Fig. 8 (c), and a 20 × 10 m outdoor park with uneven terrain, Fig. 8 (d). We compared X-Nav with the SOTA methods ViNT and NavDP. Twenty trials were conducted using random start and goal locations for each method. SR and SPL were computed and averaged across the two robots.

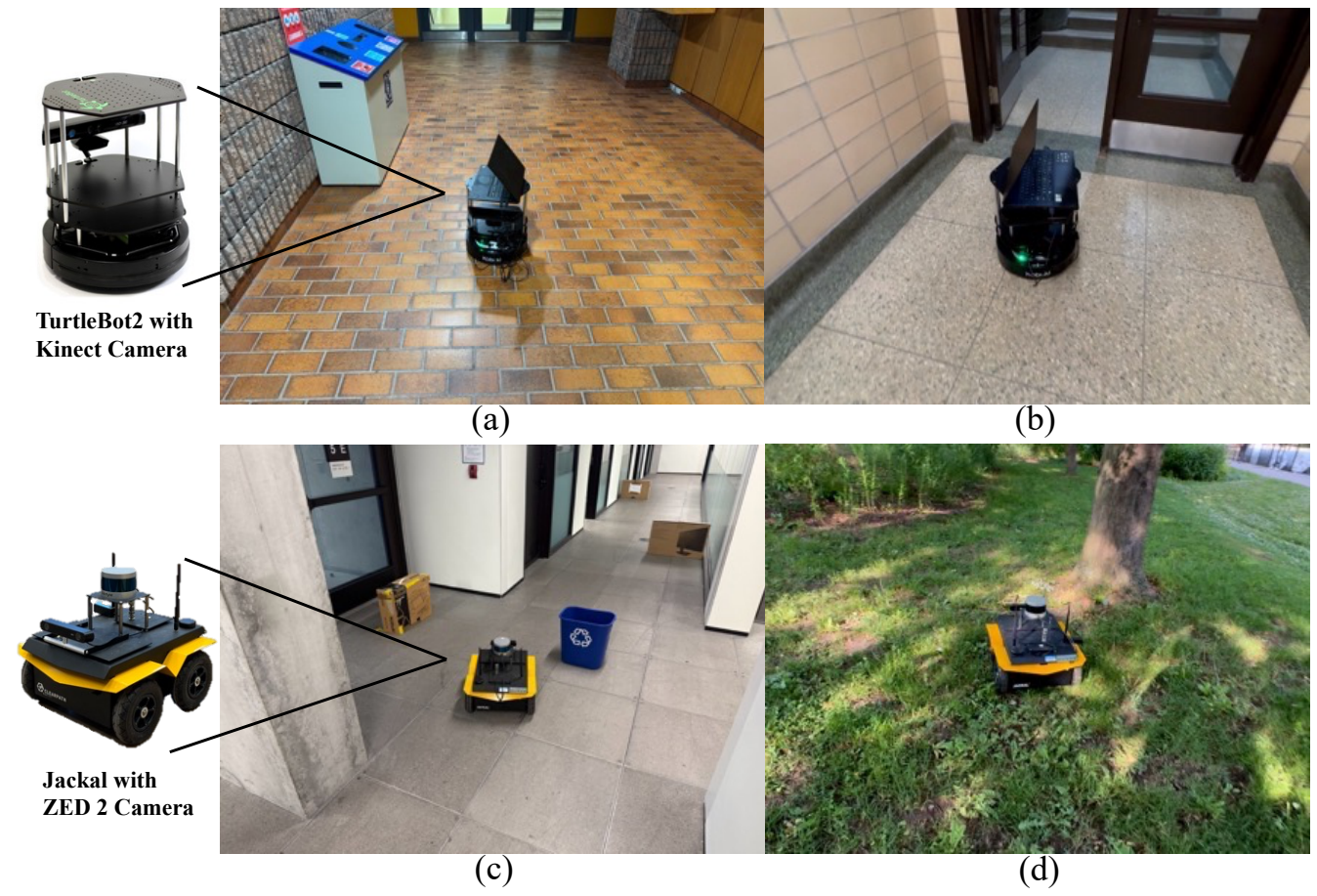

**Fig. 8.** X-Nav was deployed on the TurtleBot2 and the Jackal robots in both indoor and outdoor environments, including: (a) a hallway, (b) a doorway, (c) a corridor, and (d) a park.

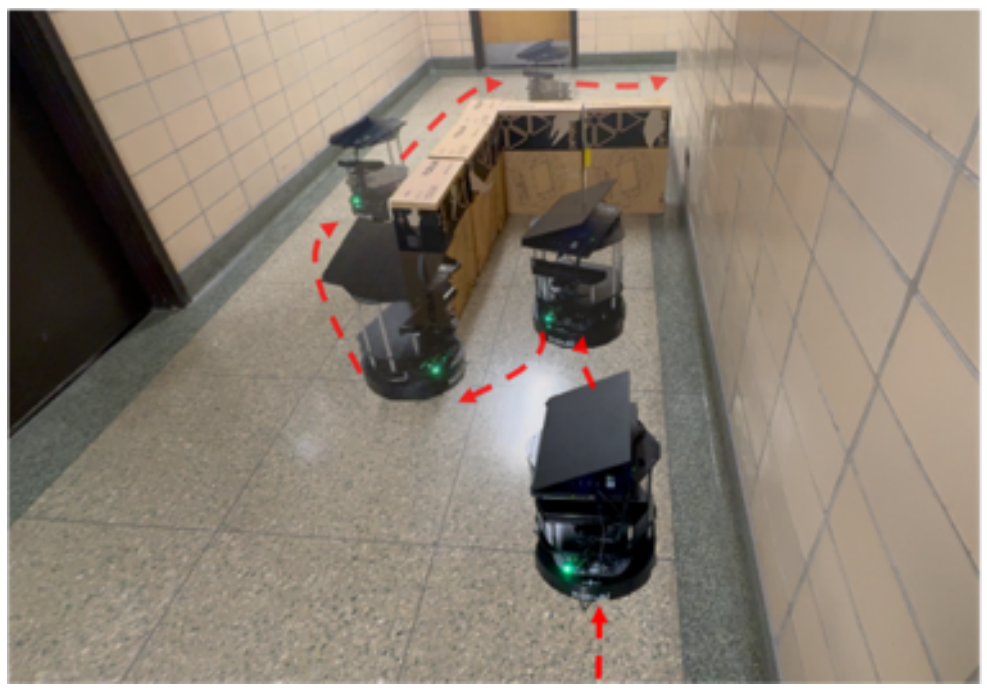

**Fig. 9.** An example of X-Nav overcoming local minima in a real-world experiment.

Table III: REAL-WORLD EXPERIMENTS

| | SR | SPL |
|---|---|---|
| ViNT | 70 | 0.63 |
| NavDP | 65 | 0.60 |
| X-Nav | 80 | 0.71 |

The results are shown in Table III. X-Nav outperformed ViNT and NavDP in both SR and SPL. These results demonstrate the zero-shot transfer capabilities of X-Nav by successfully adapting to distinct robot embodiments and camera configurations. We additionally demonstrate X-Nav's ability to avoid local minima (e.g., dead ends), Fig. 9. In this trial, the TurtleBot2 navigated a narrow corridor with an L-shaped obstacle configuration. The robot first entered the L-shape region and then made active turns to navigate out of the region and continue toward the goal. This shows that X-Nav can overcome concave obstacles with active exploration. A video of X-Nav and the supplementary material are provided at our project webpage: https://cross-embodiment-nav.github.io.

## VIII. Conclusion

In this paper, we present a novel two-stage learning framework, X-Nav, to address the problem of cross-embodiment navigation for both wheeled and quadrupedal robots. The first stage utilized deep reinforcement learning with privileged observations to train multiple expert policies. The second stage distilled these expert policies into a single general policy using a transformer model, Nav-ACT. Through extensive simulated experiments, we demonstrated the effectiveness of X-Nav in zero-shot transfer to unseen robot embodiments and photorealistic environments. A scalability study showed that X-Nav's performance scales with increasing number of random embodiments used during training. An ablation study validated our design choices and inference strategy. Real-world experiments validated the generalizability of X-Nav in both indoor and outdoor environments. Future work will extend X-Nav to more robot types (e.g., humanoids) and robot types with varying DOFs by extending the unified action space to match the maximum number of DOFs across supported embodiments. We will also incorporate object-goal navigation to expand its applicability.